\documentclass[11pt,a4paper]{article}
\usepackage[hyperref]{emnlp2020}
\usepackage{times}
\usepackage{latexsym}

\usepackage{microtype}

\aclfinalcopy 

\usepackage{hhline}
\usepackage{graphicx}
\usepackage{url}
\usepackage{multirow}
\usepackage{xcolor}
\usepackage{amsmath}
\definecolor{gainsboro}{rgb}{0.86, 0.86, 0.86}
\usepackage{stfloats}

\usepackage{fancyhdr}
\usepackage{geometry}
\usepackage{layout}
\title{KoBE: Knowledge-Based Machine Translation Evaluation}

\author{Zorik Gekhman\hspace{0.5cm}Roee Aharoni\hspace{0.5cm}Genady Beryozkin\\
\textbf{Markus Freitag}\hspace{0.5cm}\textbf{Wolfgang Macherey}\\
Google Research\\
  {\tt \{zorik,roeeaharoni\}@google.com}}
\date{}

\begin{document}
\maketitle
\begin{abstract}
We propose a simple and effective method for machine translation evaluation which does not require reference translations. Our approach is based on (1) grounding the entity mentions found in each source sentence and candidate translation against a large-scale multilingual knowledge base, and (2) measuring the recall of the grounded entities found in the candidate vs. those found in the source. Our approach achieves the highest correlation with human judgements on 9 out of the 18 language pairs from the WMT19 benchmark for evaluation without references, which is the largest number of wins for a single evaluation method on this task. On 4 language pairs, we also achieve higher correlation with human judgements than BLEU.
To foster further research, we release a dataset containing 1.8 million grounded entity mentions across 18 language pairs from the WMT19 metrics track data.
\end{abstract}

\section{Introduction}

Reliable and accessible evaluation is an important catalyst for progress in machine translation (MT) and other natural language processing tasks. While human evaluation is still considered the gold-standard when done properly \cite{laubli2020set}, automatic evaluation is a cheaper alternative that allows for rapid development cycles.
Today's prominent automatic evaluation methods like BLEU \cite{papineni-etal-2002-bleu} or METEOR \cite{banerjee-lavie-2005-meteor} rely on n-gram matching with reference translations. While these methods are widely adopted, they have notable deficiencies:
\begin{itemize}
    \item Reference translations cover a tiny fraction of all relevant input sentences or domains, and non-professional translators yield low-quality results \cite{zaidan-callison-burch-2011-crowdsourcing}.
    \item Different words in the candidate and reference translations that share an identical meaning will be penalized by simple n-gram matching, and multiple references are rarely used to alleviate this \cite{qin-specia-2015-truly}.
    \item Human translations have special traits (``Translationese'', \citealp{koppel-ordan-2011-translationese}) and reference-based metrics were shown to be biased to produce higher scores for translationese MT outputs than for valid, alternative MT outputs \cite{freitag2020bleu}.
    \item N-gram matching enables measurement of relative improvements, but does not provide an interpretable quality signal \cite{lavie2010evaluating}.
\end{itemize}

\begin{figure}[t]
    \centering
    \fcolorbox{white}{white}{\includegraphics[scale=0.465,trim={4.75cm 3.25cm 4.75cm 2.5cm},clip]{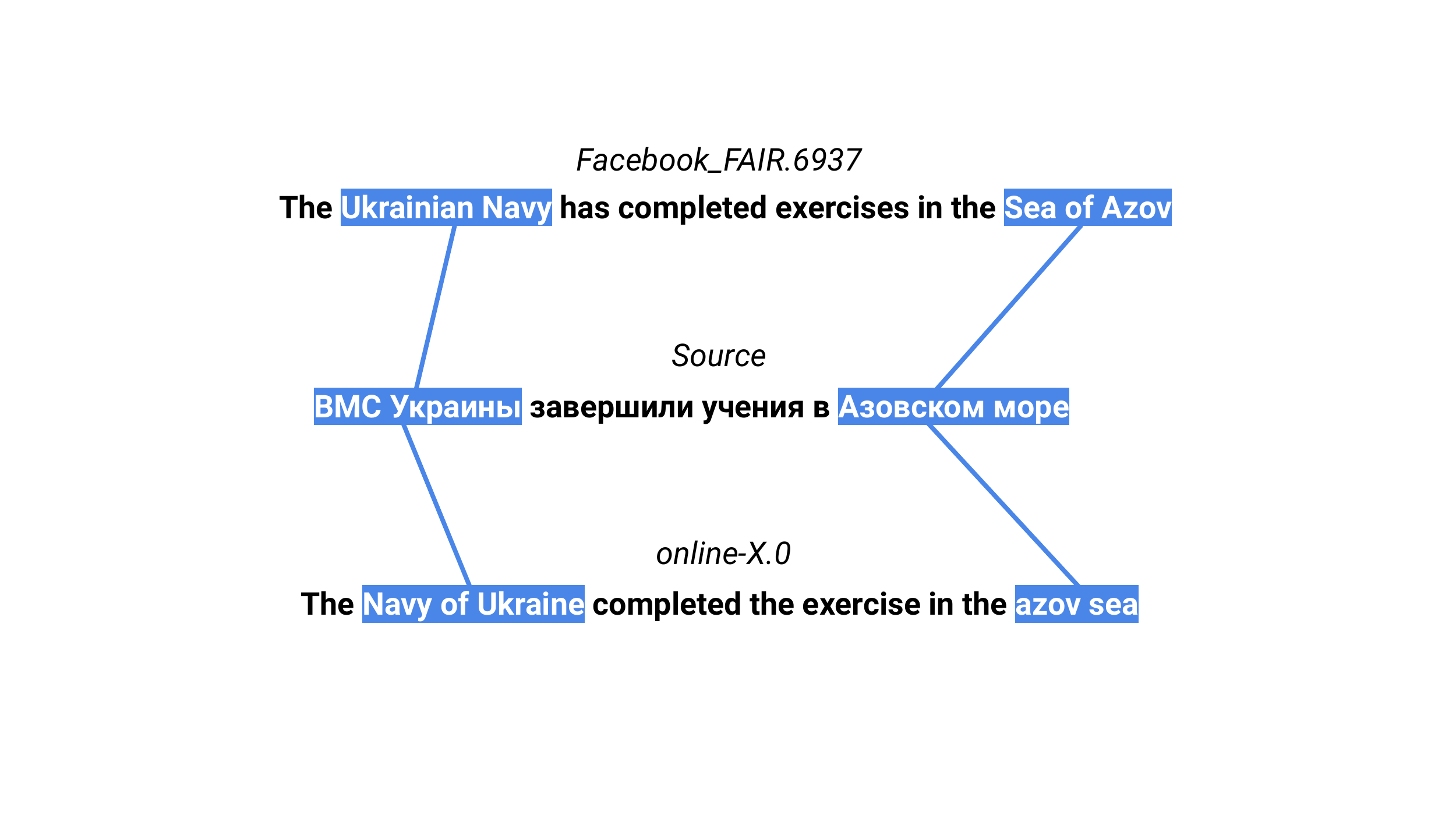}}
    \vspace{-0.6cm}
    \caption{Example annotations and entity matches using our method.}
    \label{fig:matches}
    \vspace{-0.5cm}
\end{figure}

To alleviate these issues, we propose Knowledge-Based Evaluation (KoBE), an evaluation method based on a large-scale multilingual knowledge base (KB). In our approach, we first ground each source sentence and candidate translation against the KB using entity linking \cite{mcnamee-etal-2011-cross,rao2013entity,pappu2017lightweight, gillick-etal-2019-learning, wu2019zero}. We then measure the recall for entities found in the candidate vs. entities found in the source for all sentence pairs in the test set. Matching entities are ones linked to the same KB entry in both the source and the candidate. Figure \ref{fig:matches} shows our entity matches for two candidate translations vs. the source, where different surface forms that convey the same meaning are properly matched. 

Our approach does not require reference translations, as it is based on linking entity mentions to the KB. This also makes it language-pair agnostic as long as the KB and entity linking systems cover the languages of interest. Since different words that share the same meaning should be resolved to the same entry in the KB, our method will not penalize different valid translations of the same entity. As our method measures the recall of the entities found in the source sentence, it is useful as an absolute quality signal and not just as a relative one. Finally, we can perform fine-grained error analysis using entity metadata to better understand where a system fails or succeeds in terms of entity types and domains.

To test our approach, we experiment with the ``Quality Estimation as a Metric'' benchmark (also named ``Metrics Without References'') from the WMT19 shared task on quality estimation  \cite{fonseca-etal-2019-findings}. KoBE performs better than the other participating metrics on 9 language pairs, and obtains better correlation with human raters than BLEU on 4 language pairs, even though BLEU uses reference translations and KoBE is reference-agnostic. This demonstrates that KoBE is a promising step towards MT evaluation without reference translations. 

To make our findings reproducible and useful for future work, we release the annotations we used together with scripts to reproduce our results. These entity linking annotations span over 425k sentences in 18 language pairs from 262 different MT systems, and contain 1.8 million entity mentions of 28k distinct entities.\footnote{\url{https://github.com/zorikg/KoBE}} 

To summarize, this work includes the following contributions:
\begin{itemize}
    \item We introduce KoBE, a novel knowledge-based, reference-less metric for machine translation quality estimation.
    \item We show this approach outperforms previously published results on 9 out of 18 language pairs from the WMT19 benchmark for evaluation without references.
    \item We release a data set with 1.8 million grounded entity mentions for the WMT19 benchmark to foster further research on knowledge-based evaluation.
\end{itemize}

\section{Method}

To obtain a system-level score, we first annotate all source sentences $s_i\in{S}$ and candidate translations $t_i\in{T}$ from a test set of $n$ sentence pairs using entity linking pipelines.\footnote{We used in-house systems similar to the Google Cloud Natural Language API Entity Analysis: \url{https://cloud.google.com/natural-language/docs/basics\#entity_analysis}} As a knowledge base, we used the publicly available Google Knowledge Graph Search API\footnote{\url{https://developers.google.com/knowledge-graph}} which offers entities from various domains. Unfortunately, we are not aware of any open-source multilingual KB and entity linking systems that we could rely on for the same purpose.
We then count the matches for each sentence pair; matches are all candidate entities that are linked to the same record in the KB as source entities. Entities mentioned several times are counted as individual matches, and matches are clipped by the number of appearances of each entity in the source. As a pre-processing step, we ignore entity mentions in the candidate that are not in the target language using an in-house language identification tool, which we found to improve results in early experiments.
We then compute recall by summing the number of matching entities across all sentence pairs and dividing by the the number of entities mentioned in all source sentences:
\begin{center}
$\textrm{recall} = \frac{\sum\limits_{i=0}^{n}{|\textrm{matches}(\textrm{entities}(s_i),\textrm{entities}(t_i))|}}{\sum\limits_{i=0}^{n}{|\textrm{entities}(s_i)|}}$
\end{center}

Our decision to ignore candidate entities that are not in the correct language came from an observation that for some low-resource language pairs, MT systems fail to translate the input and instead copy most of its content to the output -- see \citet{ott2018analyzing} for a similar observation. As our entity linking system is language agnostic, it was detecting the copied entities, which resulted in false matches. 

We found precision to have weaker correlation on most language pairs, as it rewards systems producing a lower number of entities -- systems that usually produced lower quality translations. Recall is more stable as the number of entities in the source is constant for all evaluated systems, and only the match count is changing. Since recall may give inflated scores when over-producing entities, we introduce an entity count penalty (ECP), inspired by BLEU's brevity penalty. ECP penalizes systems producing $c$ entities  if $c$ is more than twice the number of entities in the source,  $s$:
\begin{center}
$
    \textrm{ECP} = \left\{\begin{alignedat}{2}
        & 1 \qquad & \textrm{if} \qquad c   & < 2s \\
        & e^{(1-c/2s)} \qquad & \textrm{if} \qquad c   & \geq 2s \\
    \end{alignedat}\right.
$
\end{center}
Finally:
\begin{center}
$\textrm{KoBE} = \textrm{ECP} \cdot \textrm{recall}
$
\end{center}

\section{Experimental Setup}

The WMT conference holds a Quality Estimation track (QE) that aims to predict the quality of MT systems given the source sentences and candidate translations (without reference translations). While this was usually done at the word or sentence level, one of the novelties in WMT19 was introducing a new task for using QE as a metric at the corpus level, testing the generalization ability of QE approaches in a massive multi-system scenario \cite{fonseca-etal-2019-findings}. To test our approach, we used the same setting as in this shared task. For every language pair of the 18 evaluated pairs, we use KoBE to score the MT systems participating in same year’s news translation task. We then measure the Pearson correlation of our scores for each system with its human direct-assessment (DA) scores. To ensure a fair comparison, we recompute the correlations for the other participating metrics and confirm that we reproduce the reported scores.\footnote{More implementation details for reproducing our results are available in the supplemental material.}



\begin{table*}[t!]
\begin{center}
\begin{small}
\begin{tabular}{lccccccc}
\hline
                 & \textbf{de-en} & \textbf{fi-en} & \textbf{gu-en} & \textbf{kk-en} & \textbf{lt-en} & \textbf{ru-en} & \textbf{zh-en} \\ \hline
BLEU             & 0.849          & 0.982          & 0.834          & 0.946          & 0.961          & 0.879          & 0.899          \\ \hline
LASIM            & 0.247          & --             & --             & --             & --             & -0.31           & --             \\
LP               & -0.474          & --             & --             & --             & --             & -0.488          & --             \\
UNI              & 0.846          & \textbf{0.93}  & --             & --             & --             & 0.805          & --             \\
UNI+             & 0.85  & 0.924          & --             & --             & --             & 0.808          & --             \\
YiSi-2           & 0.796          & 0.642          & -0.566          & -0.324         & 0.442          & -0.339          & 0.94           \\
YiSi-2\_srl      & 0.804          & --             & --             & --             & --             & --             & \textbf{0.947} \\ \hline
KoBE    & \textbf{0.863} & 0.538          & \textbf{0.828} & \textbf{0.899} & \textbf{0.704} & \textbf{0.928} & 0.907          \\ \hline
\end{tabular}
\end{small}
\caption{System-level Pearson correlation with human judgements for language pairs into English from the WMT19 metrics-without-references shared task. Best QE results are marked in bold.}
\label{tab:to_en}
\vspace{0.25cm}
\begin{small}
\begin{tabular}{lcccccccc}
\hline
                 & \textbf{en-cs} & \textbf{en-de} & \textbf{en-fi} & \textbf{en-gu} & \textbf{en-kk} & \textbf{en-lt} & \textbf{en-ru} & \textbf{en-zh} \\ \hline
BLEU             & 0.897          & 0.921          & 0.969          & 0.737          & 0.852          & 0.989          & 0.986          & 0.901          \\ \hline
LASIM            & --             & 0.871          & --             & --             & --             & --             & -0.823         & --             \\
LP               & --             & -0.569         & --             & --             & --             & --             & -0.661         & --             \\
UNI              & 0.028          & 0.841          & \textbf{0.907} & --             & --             & --             & \textbf{0.919 }         & --             \\
UNI+             & --             & --             & --             & --             & --             & --             & 0.918          & --             \\
USFD             & --             & -0.224         & --             & --             & --             & --             & 0.857          & --             \\
USFD-TL          & --             & -0.091         & --             & --             & --             & --             & 0.771          & --             \\
YiSi-2           & 0.324 & 0.924          & 0.696          & \textbf{0.314} & 0.339          & \textbf{0.055} & -0.766         & -0.097         \\
YiSi-2\_srl      & --             & \textbf{0.936}          & --             & --             & --             & --             & --             & -0.118         \\ \hline
KoBE  & \textbf{0.597}          & 0.888           & 0.521          & -0.34         & \textbf{0.827}  & -0.049          & 0.895          & \textbf{0.216} \\ \hline
\end{tabular}
\end{small}
\caption{System-level Pearson correlation with human judgements for from-English language pairs from the WMT19 metrics-without-references shared task. Best QE results are marked in bold.}
\label{tab:from_en}
\end{center}
\end{table*}

\begin{table}[t!]
\begin{center}
\begin{small}
\begin{tabular}{lccc}
\hline
                 & \textbf{de-cs} & \textbf{de-fr} & \textbf{fr-de}  \\ \hline
BLEU             & 0.941          & 0.891          & 0.864           \\ \hline
ibm1-morpheme    & 0.355          & -0.509         & -0.625          \\
ibm1-pos4gram    & --             & 0.085          & \textbf{-0.478} \\
YiSi-2           & 0.606          & \textbf{0.721} & -0.53           \\ \hline
KoBE  & \textbf{0.958}   & 0.485          & -0.785          \\ \hline
\end{tabular}
\end{small}
\caption{Pearson correlation results on language pairs excluding English from the WMT19 metrics-without-references task. Best QE results are marked in bold.}
\label{tab:no_en}
\end{center}
\end{table}

\section{Results}
 We compare KoBE with all participating metrics in the shared task. We refer the reader to \citet{fonseca-etal-2019-findings} for more details about the different metrics. We also compare our results with BLEU to have a benchmark for a reference-based metric.
 
 The results for into-English language pairs are available in Table \ref{tab:to_en}. KoBE outperforms all other submissions for German-to-English, Gujarati-to-English, Kazakh-to-English, Lithuanian-to-English and Russian-to-English, making it the best system in this section in terms of the number of wins. Results for from-English language pairs are available in Table \ref{tab:from_en}. In this case KoBE outperforms the submitted systems for English-to-Czech and English-to-Kazakh with Pearson correlations of 0.597 and 0.827, and also obtains high correlations for English-to-German and English-to-Russian with 0.888 and 0.895, respectively. For English-to-Chinese we also obtain the highest correlation, but it is very low overall. Table \ref{tab:no_en} describes the results on language pairs not involving English (German-to-Czech, German-to-French and French-to-German). In this case KoBE obtains the best result for German-to-Czech with Pearson correlation of 0.958. For 4 language pairs (German-to-English,  Russian-to-English, Chinese-to-English and German-to-Czech), KoBE outperforms BLEU in terms of the correlation with human judgements. This is encouraging given that KoBE does not use reference translations while BLEU does.

 In Table \ref{tab:bleu_metric} we perform additional experiments to test whether our method can also be used as a reference-based metric, by measuring the recall of entities mentioned in the candidate translations vs. entities mentioned in the references. KoBE indeed correlates well with human judgements and outperforms BLEU on 5 out of 7 language pairs, which we find impressive given that it only considers unordered entity mentions and not on all n-grams as in BLEU. Figure \ref{fig:absolute} shows a comparison of our scores vs. BLEU and human direct-assessment on Russian-to-English. In addition to the higher correlation with human judgements (0.928 vs. 0.879), our metric produces scores which are closer to the human scores on an absolute scale.
 
 \section{Discussion and Analysis}
 Summarizing the above findings, our method obtains the best results on 9 out of 18 language pairs, which makes it the method with the largest number of wins on the WMT19 metrics-without-references benchmark. This shows that knowledge-based evaluation is a promising path towards MT evaluation without references. In comparison, the next best method is YiSi-2 \cite{lo-2019-yisi} which is based on token-level cosine-similarity using context-aware token representations from multilingual BERT \cite{devlin-etal-2019-bert}. We believe that combining our knowledge-based approach with such methods may result in even better correlation with human judgements, but leave this for future work. 
 
 As our metric is based on the recall of entities in the target with respect to the source, it is important that the entities will be properly detected in the target. A failure to detect an entity in the source will just lead KoBE to use less entities, while a failure to detect an entity in the target will lead KoBE to penalize an entity that is actually present. Our entity linking pipelines work best in English, which results in much higher correlations with human judgements when English is the target language (Table \ref{tab:to_en}) vs. the correlations when English is the source (Table \ref{tab:from_en}). We believe that as entity linking systems will improve for languages other than English, our metric will improve accordingly. 
 Another possible concern may be regarding the evaluation of sentences which do not contain any detected entities -- our analysis shows that was the case for less than 8\% of the sentences, so it did not have a large effect on the corpus-level metric.\footnote{See Figure \ref{fig:hist} in the supplemental material for a histogram of entity counts per sentence.}

\begin{figure}[t!]
\center{\includegraphics[scale=0.15,trim={0.0cm 0.0cm 0.0cm 0.0cm},clip]
{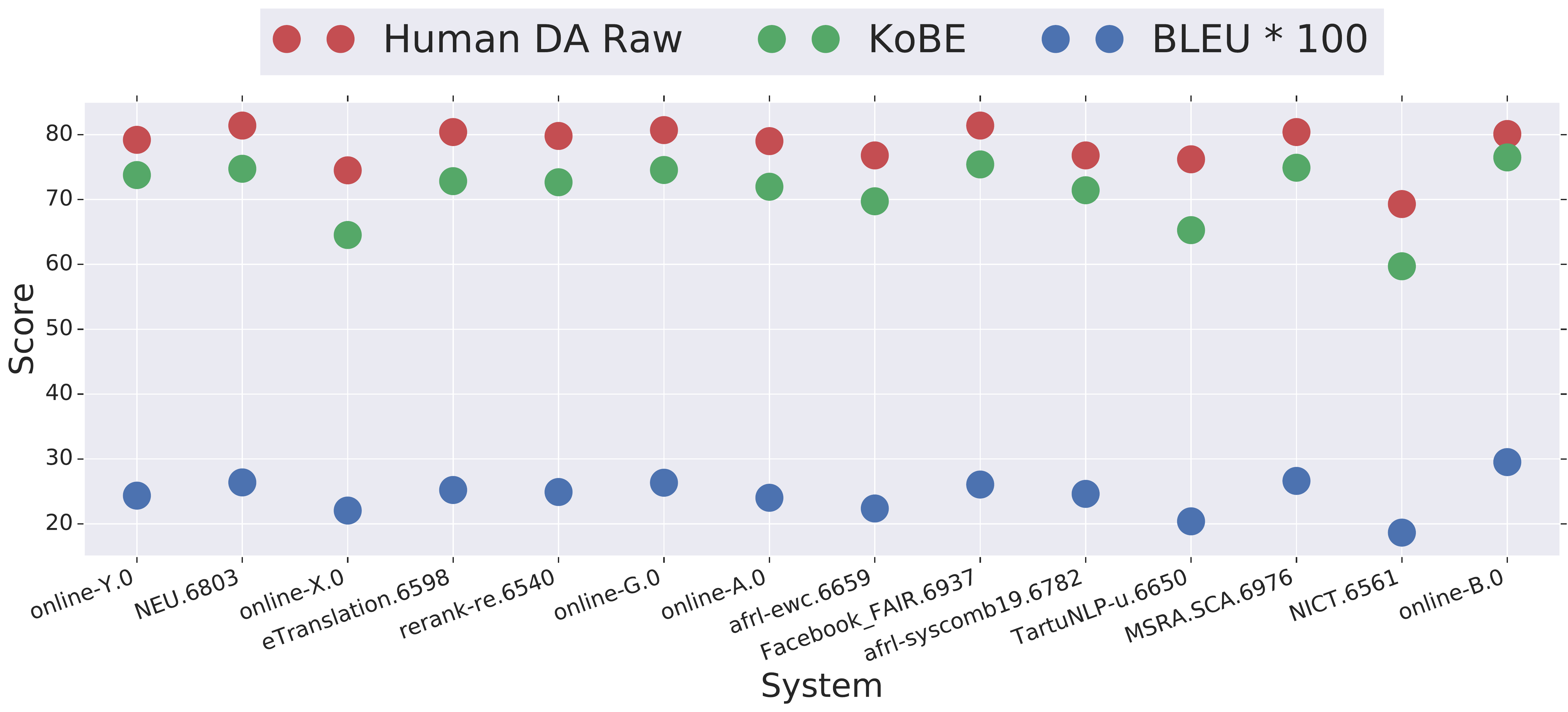}}
\vspace{-0.25cm}
\caption{Comparison of BLEU, KoBE and human judgements for Russian-to-English newstest2019. }
\label{fig:absolute}


\center{ \fcolorbox{white}{white}{\includegraphics[scale=0.345,trim={1.4cm 0.25cm 1.95cm 0.75cm},clip]
{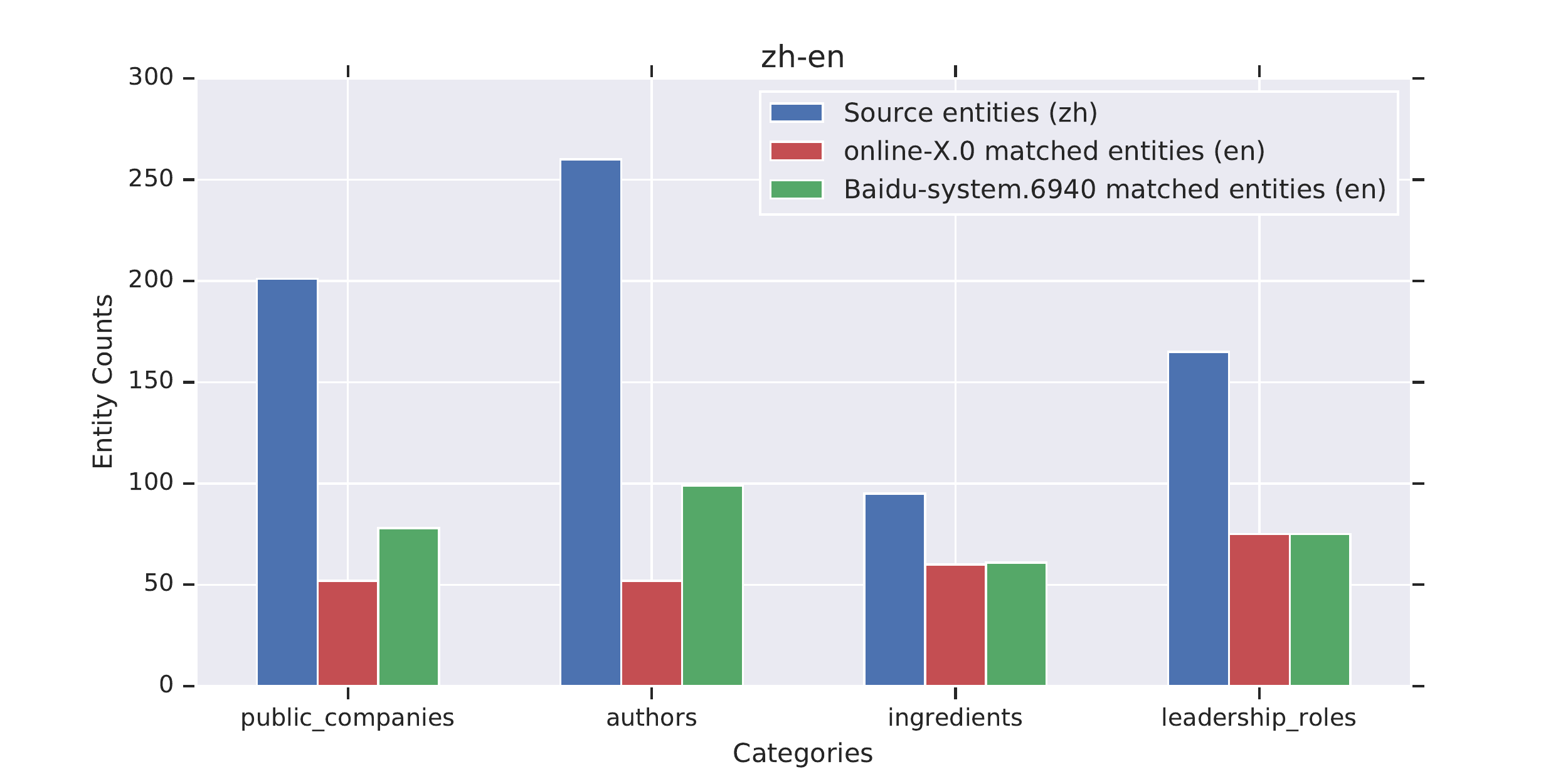}}}
\vspace{-0.75cm}
\caption{Statistics for entity categories for different Chinese-to-English MT systems and the source side.}
\label{fig:categories}
\vspace{-0.5cm}
\end{figure}

 Figure \ref{fig:categories} shows matching statistics for different MT systems across several entity categories from the KB. We can see that our scores vary across different categories between and within different systems, which can give an interpretable signal for system developers regarding where improvement efforts should be invested.

Our reproduction of the correlation results raises an issue with the current evaluation methodology in the shared task. In the published results \cite{fonseca-etal-2019-findings}, in order to support both lower-is-better metrics (e.g. TER \citealp{snover2006study}) and higher-is-better metrics (e.g. BLEU), the absolute values of the Pearson correlations are reported. However, when looking in Table \ref{tab:from_en} and Table \ref{tab:no_en} we see that the same metric may be correlated with different signs in different language pairs. This may result in wrong ranking of evaluation metrics, as the absolute value may ``cover up'' such cases. We hope future evaluations will take this detail into account. 

A possible drawback of our approach is that it only relies on entities, which do not fully cover the sentence semantics. However, in the quality estimation setting, we only have access to the source and candidate translation, which are in different languages. As different languages use different syntactic structures and vocabulary, it is hard to employ other structural cues - for example, the order of the entities may be different due to the grammatical differences between the languages. The strong correlation between our metric and human judgements shows that knowledge-based comparison is a strong indicator of translation quality in this challenging setting. This is in line with the results of \citet{freitag2020bleu} who showed that BLEU with extensively paraphrased references correlates better with human judgements than BLEU with vanilla references -- our method is ``paraphrasing'' or ``stripping down'' the candidate and reference to only contain the mentioned entities during evaluation.


\begin{table*}[t!]
\begin{center}

\end{center}
\end{table*}

\begin{table*}[ht!]
\begin{center}
\begin{tabular}{lcccccccc}
\hline
               & \textbf{de-en} & \textbf{fi-en} & \textbf{gu-en} & \textbf{kk-en} & \textbf{lt-en} & \textbf{ru-en} & \textbf{zh-en} & \textbf{mean}  \\ \hline
BLEU           & 0.849          & \textbf{0.982} & 0.834          & 0.946 & \textbf{0.961} & 0.879          & 0.899          & 0.907          \\
KoBE  & \textbf{0.906}   & 0.961          & \textbf{0.85} & \textbf{0.961}          & 0.901          & \textbf{0.954} & \textbf{0.947} & \textbf{0.926} \\ \hline
\end{tabular}
\caption{Comparison of the Pearson correlation with human judgements for BLEU and KoBE, on the into-English language pairs from the WMT19 metrics shared task. Best results are marked in bold.}
\label{tab:bleu_metric}
\end{center}
\vspace{-0.5cm}
\end{table*}

\section{Related Work}
Quality estimation for MT has been studied extensively in recent years -- see \citet{specia2018quality} for a thorough overview. Most work has been on the sentence or word level, using supervised approaches e.g. Open-Kiwi \cite{kepler-etal-2019-openkiwi}. Using semantic knowledge for MT evaluation was proposed in different approaches: METEOR \cite{denkowski-lavie-2014-meteor} used paraphrase tables for reference-based evaluation; YiSi \cite{lo-2019-yisi} and MEANT \cite{lo-2017-meant} used semantic role labeling (SRL) annotations; \citet{birch-etal-2016-hume} used the UCCA semantic annotations \cite{abend-rappoport-2013-universal} for human evaluation of MT; \citet{li-etal-2013-name} proposed a name-aware BLEU score giving more weight to named entities. \citet{babych2004comparative} conducted a comparative evaluation of named entity recognition (NER) from MT outputs, concluding that the success
rate of NER does not strongly correlate with human or automatic evaluation scores. We show contradicting results, which may stem from the better NER and MT systems available today, and from the entity linking step we add. To the best of our knowledge, our work is the first to introduce a reference-less MT evaluation method based purely on entity linking against a multilingual knowledge-base.

\section{Conclusions and Future Work}
We proposed KoBE, a method for reference-less MT evaluation using entity linking to a multilingual KB. We demonstrated the applicability of our method by achieving strong results on the WMT19 benchmark for reference-less evaluation across 9 language pairs, where in 4 cases we also outperform the reference-based BLEU. Our method is simple, interpretable and produces scores closer to human judgements on an absolute scale, while enabling more fine-grained analysis which can be useful to find weak spots in the evaluated model. 

In future work, we would like to combine knowledge-based signals with unsupervised approaches like YiSi \cite{lo-2019-yisi} and XMoverScore \cite{zhao-etal-2020-limitations} that use contextualized representations from cross-lingual LMs like multilingual BERT \cite{devlin-etal-2019-bert}. As our method does not require reference translations, we would like to explore scaling it to much larger or domain specific datasets. Other interesting directions include applying such methods to other text generation tasks like summarization or text simplification, where BLEU was shown to be problematic \cite{sulem-etal-2018-bleu}, or performing outlier-aware meta-evaluation which was recently shown to be important in such settings \cite{mathur-etal-2020-tangled}.

\section*{Acknowledgements}
We thank the reviewers for their valuable comments and suggestions for improving this work.

\bibliographystyle{acl_natbib}
\bibliography{anthology,main}

\clearpage
\appendix

\section{Supplemental Material}
\label{sec:supplemental}

\begin{table*}[b!]
\begin{center}
\begin{small}
\begin{tabular}{lccccccc}
\hline
                                  & \textbf{de-en} & \textbf{fi-en} & \textbf{gu-en} & \textbf{kk-en} & \textbf{lt-en} & \textbf{ru-en} & \textbf{zh-en} \\ \hline
sentence count                    & 2000           & 1996           & 1016           & 1000           & 1000           & 2000           & 2000           \\
source sentences with entities    & 1795           & 1672           & 796            & 751            & 934          & 1860           & 1958           \\
source entities count             & 5831           & 4645           & 1911           & 1932           & 4320           & 8230           & 15339          \\
reference entities count          & 6582           & 7070           & 3650           & 4103           & 5140           & 8413           & 18088          \\
source distinct entities count    & 2244           & 1525           & 523            & 661            & 1241           & 2404           & 3312           \\
reference distinct entities count & 2270           & 2141           & 1276           & 1329           & 1616           & 2506           & 3474           \\
common distinct entities count    & 1184           & 920            & 320            & 371            & 740            & 1446           & 1969           \\ \hline
\end{tabular}
\end{small}
\vspace{-0.25cm}
\caption{Statistics for into English language pairs from the WMT19 metrics-without-references shared task.}
\label{tab:to_en_stats}
\vspace{0.25cm}
\begin{small}
\begin{tabular}{lcccccccc}
\hline
                                  & \textbf{en-cs} & \textbf{en-de} & \textbf{en-fi} & \textbf{en-gu} & \textbf{en-kk} & \textbf{en-lt} & \textbf{en-ru} & \textbf{en-zh} \\ \hline
sentence count                    & 1997           & 1997           & 1997           & 998            & 998            & 998            & 1997           & 1997           \\
source sentences with entities    & 1870           & 1870           & 1870           & 934            & 934            & 934            & 1870           & 1870           \\
source entities count             & 9845           & 9846           & 9845           & 4711           & 4710           & 4710           & 9846           & 9845           \\
reference entities count          & 5824           & 5345           & 5113           & 2163           & 1219           & 2807           & 7563           & 10646          \\
source distinct entities count    & 3150           & 3149           & 3149           & 1941           & 1941           & 1941           & 3149           & 3149           \\
reference distinct entities count & 1446           & 1528           & 1238           & 572            & 330            & 847            & 1899           & 2739           \\
common distinct entities count    & 971            & 1006           & 899            & 364            & 202            & 555            & 1238           & 1679           \\ \hline
\end{tabular}
\end{small}
\vspace{-0.25cm}
\caption{Statistics for from-English language pairs from the WMT19 metrics-without-references shared task.}
\label{tab:from_en_stats}
\end{center}
\vspace{-0.5cm}
\end{table*}

The data used in this paper is taken from the WMT19 results.\footnote{\url{http://www.statmt.org/wmt19/results.html}} We downloaded the news translation task submissions\footnote{\url{http://data.statmt.org/wmt19/translation-task/wmt19-submitted-data-v3.tgz}} and annotated them using entity linking pipelines. We make our annotations publicly available to reproduce our results. 
We downloaded the Metrics task data\footnote{\url{http://ufallab.ms.mff.cuni.cz/~bojar/wmt19-metrics-task-package.tgz}} and obtained the submitted metrics scores, together with the standardized human direct assessment (DA) scores, from the \path{results/sys-level_scores_metrics.csv} file. We recalculated the Pearson correlations for all metrics and made sure we got the same results as reported in the WMT19 official results \cite{fonseca-etal-2019-findings}.

Our submission contains a copy of the \path{sys-level_scores_metrics.csv} file, containing the submitted metrics scores, together with the human direct assessment (DA) scores. In addition, we publish the annotations for all WMT19 news translation task submissions. The published data contains a file for each system in each language pair, as well as the annotations for the source text and reference translations. Our annotations are in json format and contain all the entities that were detected in each sentence. Each entity has an id and a start and end positions in the sentence. In addition, we publish a python script that, given the \path{sys-level_scores_metrics.csv} file and the annotations, first calculates our score for all language pairs and all systems and then calculates the Pearson correlations with human DA scores. This script and data can be used to exactly reproduce the 
results reported in the paper. We also hope that the large annotated data set will help researchers who
wish to further explore multilingual knowledge-based evaluation methods.

\begin{figure}[h!]
\center{\includegraphics[scale=0.24,trim={0.0cm 0.0cm 0.0cm 0.0cm},clip]
{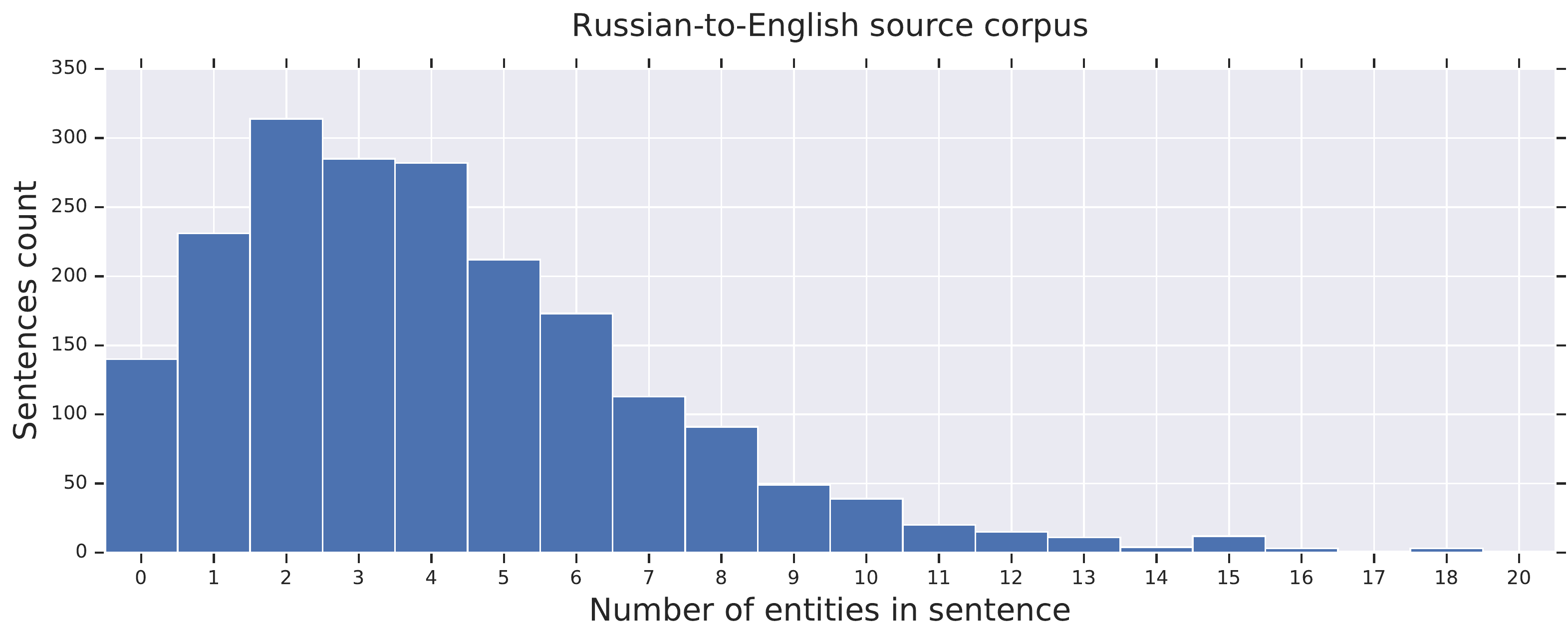}}
\vspace{-0.25cm}
\caption{Histogram of number of entities in each sentence for the Russian-to-English source corpus.}
\label{fig:hist}
\vspace{-0.5cm}
\end{figure}

We also calculate the entity statistics for each language pair using the source and the reference sentences. Table \ref{tab:to_en_stats} shows statistics for into-English language pairs, Table \ref{tab:from_en_stats} shows statistics for from-English language pairs and Table \ref{tab:no_en_stats} shows statistics for language pairs excluding English. Those tables can be also obtained by running the provided script. Note that the numbers here denote the entities that were detected by the entity linking system. Figure \ref{fig:hist} shows a histogram of entities count per sentence on the Russian-to-English source corpus.


\begin{table}[h]
\begin{center}
\begin{small}
\begin{tabular}{lccc}
\hline
                                  & \textbf{de-cs} & \textbf{de-fr} & \textbf{fr-de} \\ \hline
sentence count                    & 1997           & 1701           & 1701           \\
source sentences with entities    & 1878           & 1586           & 1634           \\
source entities count             & 8649           & 6794           & 9102           \\
reference entities count          & 5820           & 6437           & 4810           \\
source distinct entities count    & 2643           & 1571           & 1917           \\
reference distinct entities count & 1445           & 1450           & 1152           \\
common distinct entities count    & 910            & 739            & 737            \\ \hline
\end{tabular}
\end{small}
\vspace{-0.2cm}
\caption{Statistics for language pairs excluding English from the WMT19 metrics-without-references task.}
\label{tab:no_en_stats}
\end{center}
\vspace{-0.75cm}
\end{table}

\end{document}